\documentclass[letterpaper]{article} 
\usepackage{aaai2026}  
\nocopyright
\usepackage{times}  
\usepackage{helvet}  
\usepackage{courier}  
\usepackage[hyphens]{url}  
\usepackage{graphicx} 
\usepackage{natbib}  
\usepackage{caption} 
\usepackage{algorithm}
\usepackage{algorithmic}

\usepackage{newfloat}
\usepackage{listings}

\DeclareCaptionStyle{ruled}{labelfont=normalfont,labelsep=colon,strut=off} 
\floatstyle{ruled}
\newfloat{listing}{tb}{lst}{}
\floatname{listing}{Listing}
\usepackage{multirow}
\usepackage{booktabs}
\usepackage{amsfonts}
\usepackage{amsmath}

\title{Alignment-free Raw Video Demoiréing}

\author{
    Shuning Xu\textsuperscript{\rm 1}, Xina Liu, Binbin Song, Xiangyu Chen, Qiubo Chen, Jiantao Zhou \\
}
\affiliations{
    \textsuperscript{\rm 1}University of Macau \\
    yc07425@um.edu.mo
}

\usepackage{bibentry}

\begin{document}

\maketitle

\begin{abstract}
Video demoiréing aims to remove undesirable interference patterns that arise during the capture of screen content, restoring artifact-free frames while maintaining temporal consistency. Existing video demoiréing methods typically utilize carefully designed alignment modules to estimate inter-frame motion for leveraging temporal information; however, these modules are often complex and computationally demanding. Meanwhile, recent works indicate that using raw data as input significantly enhances demoiréing performance. Building on this insight, this paper introduces a novel alignment-free raw video demoiréing network with frequency-assisted spatio-temporal Mamba (DemMamba). It incorporates sequentially arranged Spatial Mamba Blocks (SMB) and Temporal Mamba Blocks (TMB) to effectively model the inter- and intra-relationships in raw video demoiréing. The SMB employs a multi-directional scanning mechanism coupled with a learnable frequency compressor to effectively differentiate interference patterns across various orientations and frequencies, resulting in reduced artifacts, sharper edges, and faithful texture reconstruction. Concurrently, the TMB enhances temporal consistency by performing bidirectional scanning across the temporal sequences and integrating channel attention techniques, facilitating improved temporal information fusion. Extensive experiments demonstrate that DemMamba surpasses state-of-the-art methods by 1.6 dB in PSNR, and also delivers a satisfactory visual experience.

\end{abstract}

\section{Introduction}
\label{sec:intro}

\begin{figure}[!htbp]
\centering
\includegraphics[width=1\linewidth]{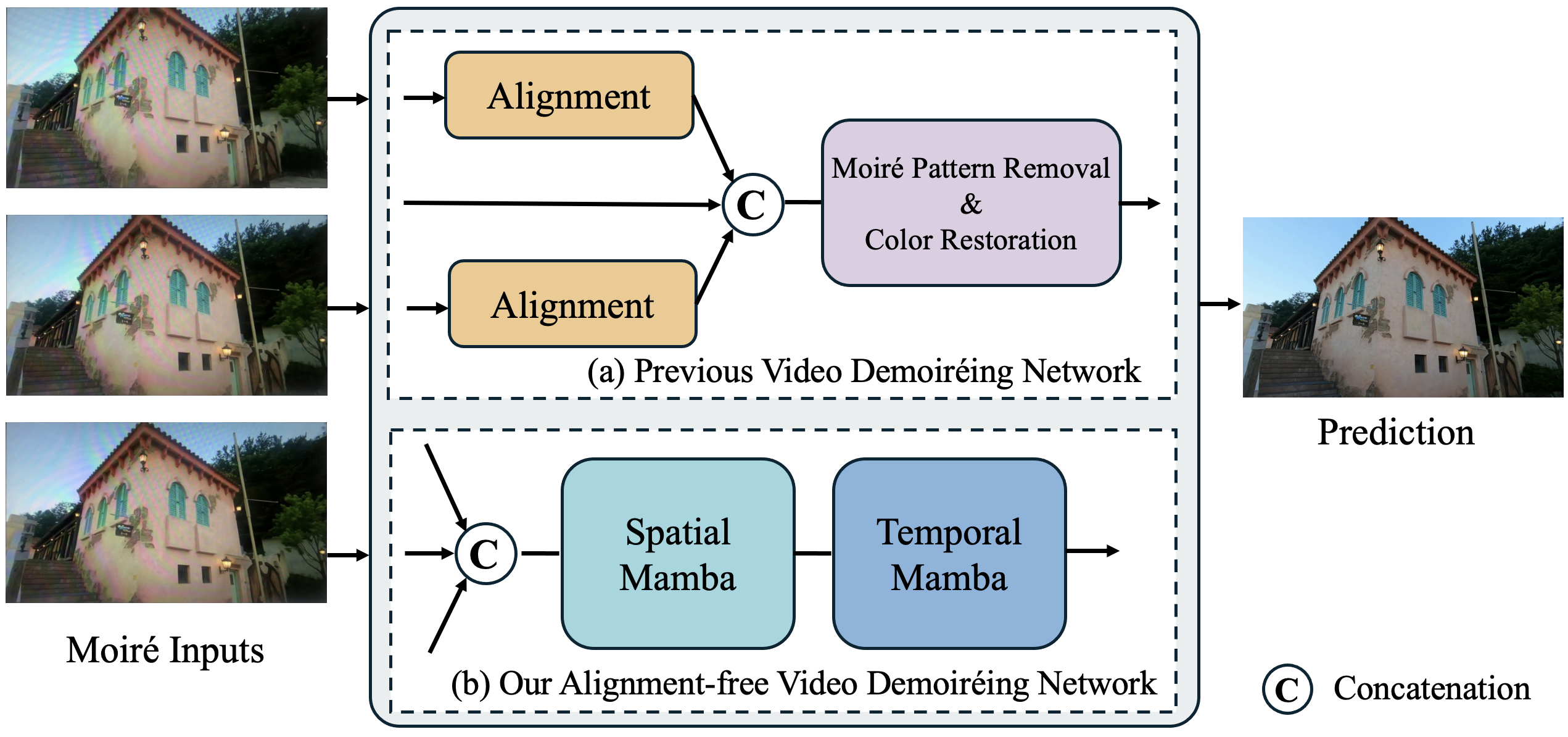}
\caption{Motivation of our alignment-free raw video demoiréing network. 
(a) Previous methods utilize a well-designed alignment module. (\textit{Computational unfriendly.})
(b) We propose Spatial Mamba and Temporal Mamba for inter- and intra-relationship modeling. (\textit{Effective and efficient.})
}
\label{fig:motivation}
\end{figure}

Moiré patterns appear as repetitive, wave-like, and colorful distortionsin images or videos captured from screens. These patterns result from frequency aliasing between the color filter array of camera sensors and the LCD screen subpixels~\cite{wang2023coarse}, resulting in a degraded visual experience. Image demoiréing has received increasing attention from the research community~\cite{cheng2019multi, he2020fhde, zheng2020image, wang2021image, niu2023progressive, wang2023coarse}. In these image demoiréing methods, the restoration process can be modeled as a combination of moire pattern removal and color correction. Many works indicate that addressing moiré patterns early in the raw-data pipeline yields superior results compared to corrections performed post-Image Signal Processor (ISP) in the sRGB domain~\cite{yue2022recaptured, cheng2024recaptured, xu2024image}, as the raw domain is free of nonlinear processes (e.g., gamma correction and demosaicing) that exacerbate moiré artifacts. Given the increasing availability of raw data in smartphones and DSLR cameras, demoiréing in the raw domain is both practical and beneficial.

With the growing demand for video capture, several learning-based methods have been developed to mitigate undesirable moiré patterns in videos~\cite{dai2022video, quan2023deep, xu2024direction, liu2024video, niu2024std, cheng2024recaptured}. However, video demoiréing presents greater complexities compared to image demoiréing. It demands not only the restoration of high-quality, moiré-free frames with natural colors but also significant effort to maintain temporal consistency.

To leverage information from adjacent frames and maintain temporal consistency, most of the existing video demoiréing methods rely heavily on sophisticated alignment modules, such as the pyramid cascading deformable (PCD) module proposed by EDVR~\cite{wang2019edvr}. This module employs deformable convolutions to align multiple frames, as depicted in Fig.~\ref{fig:motivation}(a). Although this approach appears to be straightforward, it incurs a high computational burden, limiting its efficiency, especially for processing high-resolution, long-duration videos.

Recently, State Space Models (SSMs)~\cite{kalman1960new}, originating from control system theory, have demonstrated advantages in natural language processing\cite{gu2021efficiently, gu2023mamba} and computer vision~\cite{zhu2024vision, shi2024vmambair}, primarily due to their linear complexity in processing long sequences. In low-level vision tasks, some works have successfully adapted the Mamba architecture for image restoration~\cite{guo2025mambair, li2024fouriermamba, deng2024cu, bai2024retinexmamba}, effectively balancing global receptive fields and computational efficiency. This prompts us to further delve into the potential of Mamba in video restoration, particularly targeting efficient removal of moiré patterns from raw videos while preserving temporal consistency. To facilitate modeling effective inter-frame and intra-frame relationship in raw videos with moiré patterns, as demonstrated in Fig.~\ref{fig:motivation}(b), we propose the sequentially arranged Spatial Mamba and Temporal Mamba blocks.

In this paper, we propose an alignment-free raw video demoiréing network, DemMamba, featuring frequency-assisted spatio-temporal Mamba.
Our principal design employs sequentially arranged Spatial Mamba Blocks (SMB) and Temporal Mamba Blocks (TMB) to model inter- and intra-relationships, thereby efficiently removing moiré patterns while maintaining temporal consistency.
Considering that the orientation of moiré patterns can vary substantially even within a single image, we introduce a multi-directional scanning mechanism within the SMB. This mechanism minimizes directional bias, accommodates diverse textural interferences, and enhances moiré artifact suppression. Additionally, we integrate a learnable frequency-domain compressor within the SMB to selectively attenuate spectral components associated with pronounced moiré patterns.
For TMB, the bidirectional scanning effectively models inter-frame relationships, enhancing the utilization of neighboring information. Furthermore, we integrate a channel attention block within the TMB to reinforce inter-channel dependencies, thereby further suppressing moiré patterns and ensuring robust temporal consistency throughout the video sequence.

In summary, our contributions are listed as follows:
\begin{itemize}
\item[$\bullet$] We propose an alignment-free framework for raw video demoiréing. It alternately applies Spatial Mamba and Temporal Mamba blocks to model intra- and inter-frame relationships, effectively removing moiré patterns while preserving temporal consistency.

\item[$\bullet$] For raw video demoiréing, we propose the SMB embedded with a multi-direction scanning mechanism and a learnable frequency-domain compressor to suppress prominent moiré patterns. Also, we introduce the TMB to model inter-frame relationships, exploiting neighboring information and ensuring temporal consistency.
\item[$\bullet$] Extensive experiments are conducted on both raw video and image demoiréing datasets. DemMamba surpasses state-of-the-art methods by 1.6 dB in PSNR, and also delivers a visually appealing result.
\end{itemize}

\section{Related Works}
\label{sec:related}
\subsection{Image and Video Demoiréing}
Moiré patterns result from the interference between two similar frequency patterns, frequently appearing in screen captures and severely degrading the visual experience.
Several learning-based solutions have been proposed to eliminate moiré patterns from images~\cite{liu2018demoir, he2019mop, liu2020mmdm, liu2020self, zhang2023real, xiao2024p, yang2025dsdnet}.
Compared to image demoiréing, video demoiréing~\cite{dai2022video, quan2023deep, xu2024direction, niu2024std} presents a greater challenge, as it requires generating temporally consistent moiré-free predictions.
VDMoiré~\cite{dai2022video} introduces a straightforward video demoiréing model that leverages implicit feature space alignment and selective feature aggregation. Additionally, it incorporates a novel relation-based consistency loss to enhance temporal consistency across frames.
DTNet~\cite{xu2024direction} proposes a direction-aware, temporally guided bilateral learning network for video demoiréing, utilizing multi-scale alignment and deformable convolution during the alignment process.
RawVD~\cite{cheng2024recaptured} employs a pyramid cascading deformable module for the alignment process and utilizes a dual-branch convolution approach that combines cross-channel and color-group convolutions to effectively remove moiré patterns and enhance visual structures.
Notably, existing video demoiréing methods rely on complex alignment modules, increasing computational complexity. Therefore, we introduce DemMamba, an alignment-free method designed to effectively and efficiently eliminates moiré patterns from videos, maintaining temporal consistency while reducing computational demands.

\subsection{Deep Raw Image and Video Restoration}
Raw pixel data inherently offer richer information for visual tasks, owing to their greater bit depth and extended radiance range.
Recently, several low-level vision tasks have adopted raw data, including low-light enhancement~\cite{huang2022towards, dong2022abandoning, jin2023dnf}, super-resolution~\cite{xing2021end, yue2022real, luo2024and}, reflection removal~\cite{lei2021robust, song2023real}, and moiré pattern removal~\cite{yue2022recaptured, xu2024image, cheng2024recaptured}.
RRID~\cite{xu2024image} proposes an image demoiréing network that employs paired raw-sRGB data to facilitate the color recovery process.
RawVD~\cite{cheng2024recaptured}, the first work to explore raw video demoiréing, constructs the RawVDemoiré dataset.
In this paper, considering that raw pixels offer more information and that moiré patterns are less apparent in the raw domain, we perform video demoiréing in the raw domain.

\subsection{State Space Models}
State Space Models (SSMs)~\cite{kalman1960new} have garnered significant attention recently due to their notable efficiency in employing state space transformations to manage long-term dependencies in language sequences~\cite{gu2021combining}.
More recently, Mamba~\cite{gu2023mamba} is designed with a selective scan mechanism and an efficient hardware architecture, demonstrating substantial performance improvements.
These advancements validate the effectiveness of Mamba in visual tasks, extending its applications to semantic segmentation~\cite{liu2024cm, zhu2024rethinking, wan2024sigma}, image classification~\cite{liu2024vmamba, zhu2024vision, patro2024simba, yang2024plainmamba, behrouz2024mambamixer}, video understanding~\cite{wang2023selective, chen2024video}, and image restoration~\cite{guo2025mambair, zhen2024freqmamba, wu2024rainmamba, li2024fouriermamba, lin2024pixmamba, kong2024efficient}.
Building on Mamba’s success, this paper explores the potential of Mamba in raw video demoiréing.

\section{Proposed Method}
\label{sec:method}
\subsection{Motivation}
\begin{figure}[!ht]
\centering
\includegraphics[width=1\linewidth]{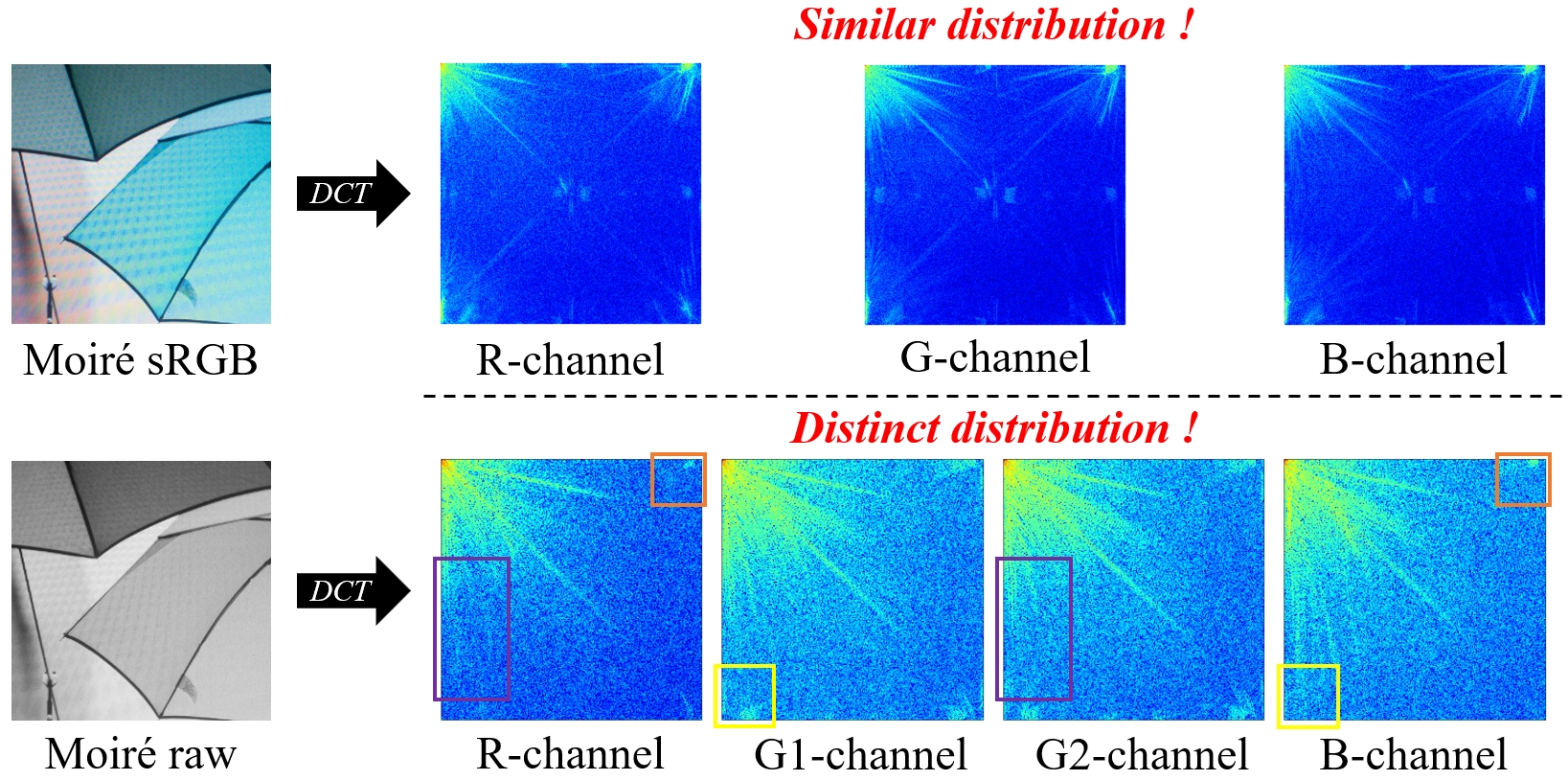}
\caption{DCT analysis of the moiré raw and sRGB images.}
\label{fig:dct}
\end{figure}

\textbf{Utilizing raw data.}
We advocate using raw data for video demoiréing, as moiré patterns are more distinguishable from content in the raw domain because the raw data remains unaffected by nonlinear operations in the ISP~\cite{cheng2024recaptured}. This distinction is clearly illustrated in Fig.\ref{fig:dct}, where the Discrete Cosine Transform (DCT) is applied to images containing moiré patterns in both raw and sRGB domains.
The raw domain exhibits distinct frequency distributions across color channels, whereas the DCT spectra for the R, G, and B channels in the sRGB domain show higher similarity, making moiré patterns less discernible.
Therefore, we prefer raw data over sRGB data for video demoiréing.

\noindent
\textbf{Applying Mamba-based framework.}
Mamba~\cite{gu2023mamba} incorporates a selective scan mechanism and exhibits linear complexity when processing long sequences, making it intuitively suitable for video restoration tasks. Building upon the foundation established by Mamba, we introduce a novel configuration of Spatial Mamba Blocks (SMBs) and Temporal Mamba Blocks (TMBs), designed to effectively model both inter-frame and intra-frame relationships for video demoiréing. SMB is designed to produce high-quality, moiré-free video frames with natural color fidelity. Meanwhile, TMB leverage temporal data from adjacent frames to remove moiré patterns while preserving temporal consistency across the video sequence.

\begin{figure*}[!htbp]
\centering
\includegraphics[width=1\linewidth]{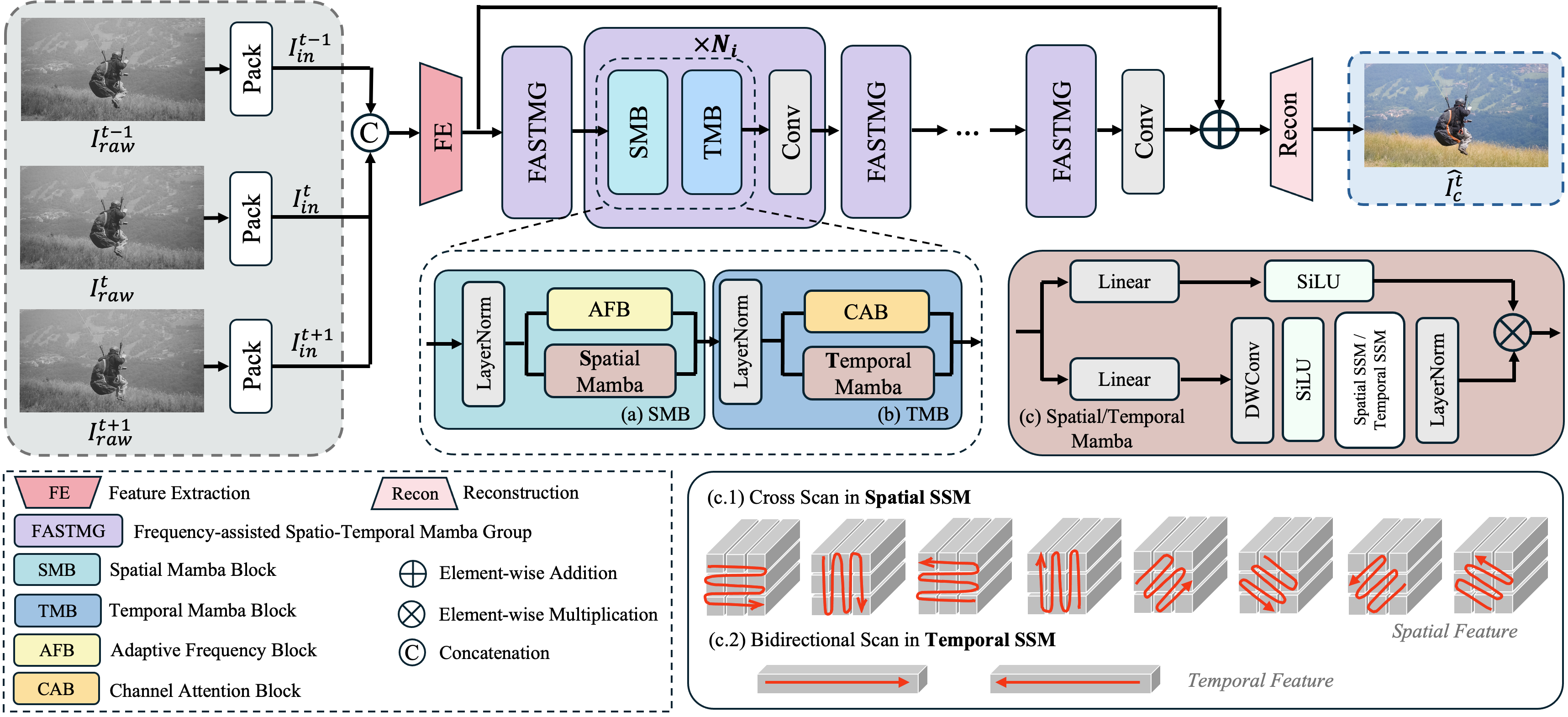}
\caption{The overview and detailed structures of our proposed DemMamba.
}
\label{fig:architecture}
\end{figure*}

\subsection{Architecture}
\label{sec:arch}
Fig.~\ref{fig:architecture} demonstrates an overview of the proposed DemMamba for raw video demoiréing.
Given three consecutive raw moiré frames $\{\mathbf{I}^{t-1}_{raw}, \mathbf{I}^{t}_{raw}, \mathbf{I}^{t+1}_{raw}\}$, the objective of raw video demoiréing is to restore the reference frame $\mathbf{I}^{t}_{raw}$ to a clean output in the sRGB domain, denoted as $\mathbf{\hat{I}}^t_c$.
The complete pipeline includes preprocessing, feature extraction, a series of Mamba groups, and a reconstruction module.

For the preprocessing, the raw moiré images $\mathbf{I}^{t+i}_{raw} \in \mathbb{R}^{H\times W \times 3}$, $i\in \{{-}1, 0, {+}1\}$, where $H \times W$ represents the spatial resolution, are initially packed into the 4-channel RGGB format $\mathbf{I}^{t+i}_{in} \in \mathbb{R}^{\frac{H}{2}\times \frac{W}{2} \times 4}$.
To address various moiré patterns across frames, we concatenate the packed inputs $\mathbf{I}^{t+i}_{in}$, and perform shallow feature extraction to obtain the shallow features $\mathbf{F}_0\in\mathbb{R}^{\frac{H}{4}\times \frac{W}{4}\times C}$ through a single convolutional layer with a stride of two , where $C$ is the number of feature channels.
Subsequently, we employ several Frequency-assisted Spatio-Temporal Mamba Groups (FASTMG) to learn the spatio-temporal features. Within each FASTMG, spatial and temporal dependencies are handled by specific modules. Each FASTMG consists of $M$ SMBs and $M$ TMBs organized sequentially to model both inter- and intra-frame relationships effectively, followed by a convolutional layer.
In SMB, a multi-direction scan is employed to capture spatial features in the Spatial Mamba, providing both a global effective receptive field and linear computational complexity relative to the input size. In conjunction with the Spatial Mamba, we introduce an Adaptive Frequency Block (AFB) with an adaptive compressor in the frequency domain to attenuate the specific frequencies of moiré patterns.
TMB processes flattened temporal sequences through simultaneous forward and backward SSMs, enhancing its capacity for temporal-aware processing. A Channel Attention Block (CAB) is embedded to further enhance temporal information interactions by exploiting inter-channel relationships among features.
To preserve and integrate shallow and spatio-temporal features, we incorporate a global residual connection. Finally, the reconstruction module derives $\mathbf{\hat{I}}^t_c$. The reconstruction module consists of several convolutional layers and a sub-pixel convolution layer to upsample the features, thus reconstructing the predicted clean image with height $H$ and width $W$.

\begin{figure}[!htbp]
\centering
\includegraphics[width=1\linewidth]{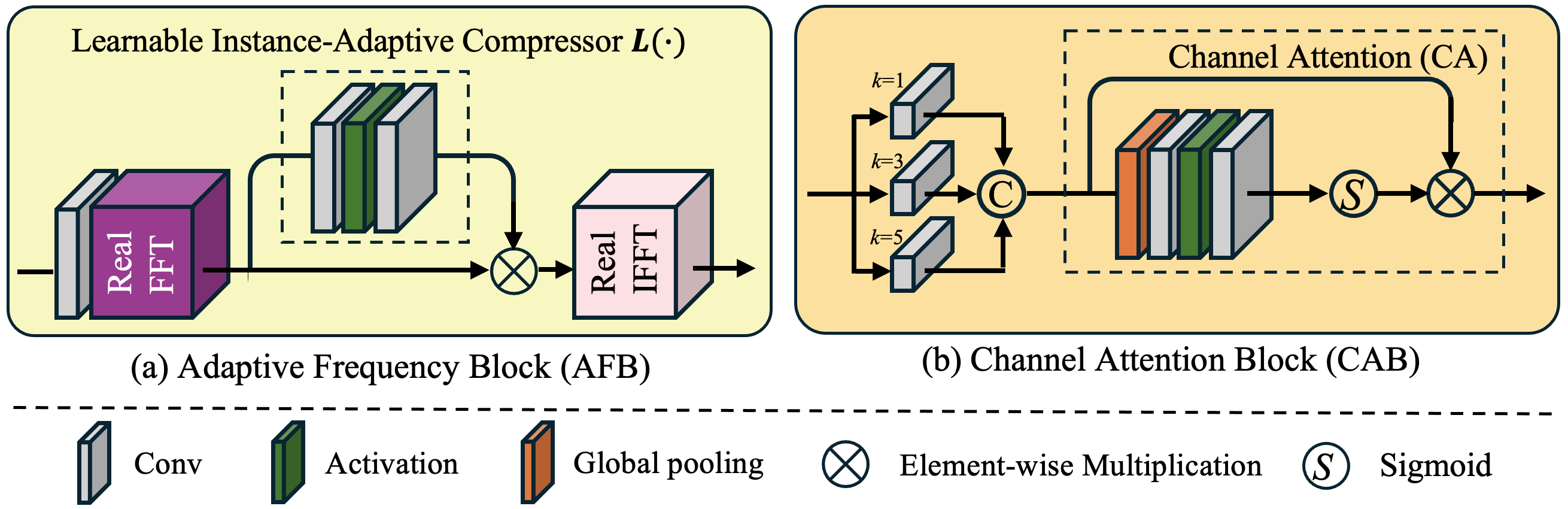}
\caption{The structure of (a) AFB and (b) CAB.}
\label{fig:CABAFB}
\end{figure}

\subsubsection{Design of SMB}
\label{sec:smb}
SMB, depicted in Fig.~\ref{fig:architecture}(a), is designed to establish spatial relationships among moiré patterns, facilitating the production of high-quality, moiré-free video frames with  authentic color fidelity.
Inspired by the success of Mamba in long-range modeling with linear complexity, we introduce the Spatial Mamba, shown in Fig.~\ref{fig:architecture}(c), to capture long-range dependencies in the spatial domain.

Following the Vision SSM design~\cite{liu2024vmamba}, the input feature $\mathbf{X} \in \mathbb{R}^{H \times W \times C}$ is processed through two parallel branches.
In the first branch, the feature channels are expanded to $\lambda C$ by a linear layer, where $\lambda$ is a pre-defined channel expansion factor, followed by depth-wise convolution, the SiLU activation function, the 2D-SSM layer, and LayerNorm. In the second branch, the feature channels are similarly expanded to $\lambda C$ using a linear layer, then processed with the SiLU activation function. 
Afterward, features from both branches are aggregated using the Hadamard product. 
The final output of the Spatial Mamba $\mathbf{X}_{SM}$ is obtained by projecting the channel number back to $C$. The process in Spatial Mamba can be described as:
\begin{equation}
\begin{aligned}
&\mathbf{X}_1 = \mathrm{LN(2D\text{-}SSM(SiLU(DWConv(Linear}(\mathbf{X}))))),\\
&\mathbf{X}_2 = \mathrm{SiLU(Linear}(\mathbf{X})),\\
&\mathbf{X}_{SM} = \mathrm{Linear}(\mathbf{X}_1 \odot \mathbf{X}_2),
\end{aligned}
\end{equation}
where $\mathbf{X}_1$, $\mathbf{X}_2$ are the results from the two branches, $\mathrm{DWConv}$ denotes depth-wise convolution, and $\odot$ is the Hadamard product. 

To extend selective SSM to 2D spatial data, we introduce a 2D Selective Scan Module for multi-direction scanning in Spatial SSM. 
Motivated by the fact that moire patterns often exhibit diverse directional characteristics, conventional single-directional scanning methods are insufficient for accurately modeling and removing these complex artifacts. 
By scanning along multiple orientations, the module can comprehensively capture moiré patterns that appear at arbitrary angles, ensuring robust detection of interference edges regardless of their direction.
As shown in Fig.~\ref{fig:architecture}(c.1), the 2D image feature is flattened and scanned along eight directions. Long-range dependencies are captured in each sequence using discrete state-space equations, and the sequences are summed and reshaped to restore the 2D structure.

Considering that moiré patterns arise from frequency aliasing, we propose AFB to aid in the removal of moiré patterns in the frequency domain, operating in parallel with the Spatial Mamba.
It is assumed that the frequency spectrum of moiré patterns tends to be relatively consistent within small patches~\cite{xu2024direction}. In signal processing, a band-reject filter allows most frequencies to pass unaltered while attenuating those within a specific range to very low levels. Given that different patches may require the attenuation of distinct frequencies, identifying a frequency prior for each moiré image patch is time-consuming. Consequently, as shown in Fig.~\ref{fig:CABAFB}(a), we construct a learnable instance-adaptive compressor $L(\cdot)$ in AFB, composed of several convolutional layers, to dynamically target and filter frequencies associated with moiré patterns while preserving the original image content.
We use Real FFT and its inverse process to apply $L(\cdot)$ in the frequency domain as:
\begin{equation}
\begin{aligned}
&\mathbf{X}_F = F(\mathbf{X}),\\
&\hat{\mathbf{X}} = F^{-1}((L(\mathbf{X}_F))\odot \mathbf{X}_F),
\end{aligned}
\end{equation}
where $F(\cdot)$ and $F^{-1}(\cdot)$ represent the Real FFT and its inverse process, respectively.

Finally, the output of the SMB is produced through a weighted summation of the outputs from the Spatial Mamba and the AFB as
$ \mathbf{X}_{SMB} = \mathbf{X}_{SM} + \alpha_1  \mathbf{X}_F$, where $\alpha_1$ is a pre-defined weighting coefficient that enables the model to effectively reduce moiré patterns while maintaining overall visual quality and detail richness.

\begin{table*}[tbp]
    \footnotesize
    \centering
    \begin{tabular}{@{}ccccccccc@{}}
        \toprule
        & \textbf{Method} & \textbf{PSNR}$\uparrow$ & \textbf{SSIM}$\uparrow$ & \textbf{LPIPS}$\downarrow$ & \textbf{FVD$\downarrow$} & \textbf{FSIM$\uparrow$} & \textbf{Parameters (M)} & \textbf{Runtime (s)} \\
        \midrule
        \multirow{2}{*}{\textbf{Image}} 
        & RDNet & 25.892 & 0.8939 & 0.1508 & 553.50 & 0.9130 & \underline{2.722} & 2.514 \\
        & RRID & 27.283 & 0.9029 & 0.1168 & 311.85 & 0.9306 & \textbf{2.374} & \textbf{0.501} \\
        \midrule
        \multirow{7}{*}{\textbf{Video}} 
        & VDMoiré & 27.277 & 0.9071 & 0.1044 & 405.22 & 0.9158 & 5.836 & 1.057 \\
        & VDMoiré$\ast$ & 27.747 & 0.9116 & 0.0995 & 395.01 & 0.9155 & 5.838 & 1.125 \\
        & DTNet & 27.363 & 0.8963 & 0.1425 & 573.34  & 0.9174 & 3.987 & 0.972 \\
        & DTNet$\ast$ & 27.892 & 0.9055 & 0.1135 & 433.39 & 0.9257 & 3.368 & 1.050 \\
        & RawVD & 28.706 & \textbf{0.9201} & \underline{0.0904} & 355.75 & 0.9246 & 6.585 & 1.247 \\
        & Ours & \underline{30.069} & 0.9192 & 0.1115 & \underline{289.48} & \underline{0.9350} & 2.941 & \underline{0.592} \\
        & Ours-L & \textbf{30.317} & \underline{0.9195} & \textbf{0.0900} & \textbf{276.29} & \textbf{0.9361} & 4.382 & 0.686 \\
        \bottomrule
    \end{tabular}
    \caption{Comparison with state-of-the-art image and video demoiréing methods for raw video demoiréing in terms of PSNR, SSIM, LPIPS, FVD, FSIM and computing complexity. The best result is highlighted in bold and the second best is underlined.}
    \label{table:nips23}
\end{table*}

\subsubsection{Design of TMB}
\label{sec:tmb}
TMB, depicted in Fig.~\ref{fig:architecture}(b), is designed to learn inter-frame relationships that enhance the utilization of neighboring information, thereby alleviating moire patterns and preserving temporal consistency throughout the video sequence. 
Serving as the core structure of TMB, the Temporal Mamba is analogous to the aforementioned Spatial Mamba but employs a different scanning strategy.
Specifically, we opt for forward and backward scanning~\cite{zhu2024vision} on the flattened temporal sequences, given that the temporal axis is one-dimensional, as illustrated in Fig.~\ref{fig:architecture}(c.2).
This bidirectional scanning effective models the interrelations across the temporal domain.

To enhance temporal information interactions, we introduce a CAB to exploit the inter-channel relationship among features. As shown in Fig.~\ref{fig:CABAFB}(b), CAB consists of a set of convolutional layers with different kernel size ($k=1,3,5$) and a channel attention (CA) module. We utilize convolutional kernels of varying sizes to capture temporal dependencies across diverse temporal length. This parallel architecture enables the network to learn the most effective kernel configurations, thereby enhancing its ability to model complex temporal patterns. Subsequently, a standard CA module~\cite{zhang2018image} adaptively rescales channel-wise features. Finally, the output of the TMB can be computed by the weighted summation of the output from Temporal Mamba $\mathbf{X}_{TM}$ and $\mathbf{X}_{CAB}$ as $ \mathbf{X}_{TMB} = \mathbf{X}_{TM} + \alpha_2  \mathbf{X}_{CAB} $, where $\alpha_2$ serves as the pre-defined weighting coefficient in TMB that enables the model to effectively integrate temporal dependencies while maintaining overall feature coherence.

\subsection{Loss Function}
We train our framework in an end-to-end manner with the overall training objective defined as:
\begin{equation}
\label{eq:loss}
\mathcal{L} = \|\hat{\mathbf{I}}^t_c - \mathbf{I}^t_{c}\|_1 + \| \Phi_l(\hat{\mathbf{I}}^t_c) - \Phi_l(\mathbf{I}^t_{c}) \|_1,
\end{equation}
where $\mathbf{I}^t_{c}$ represents the ground-truth moiré-free image of frame $t$.
We employ $L_1$ loss alongside the perceptual loss~\cite{johnson2016perceptual}, which can closely reflect the human visual system's perception of image quality. $\Phi_l(\cdot)$ denotes a set of VGG-16 layers.

\section{Experiments}
\label{sec:exp}
\subsection{Experimental Setup}
\textbf{Dataset.}
We assess the effectiveness of our proposed DemMamba using the RawVDemoiré dataset~\cite{cheng2024recaptured}, which consists of 300 clean raw source videos along with their corresponding moiréd versions in the sRGB domain. RawVDemoiré dataset is divided into 250 videos for training and 50 videos for testing. Each video comprises 60 frames at a resolution of 720p (1080×720). Four camera-screen combinations are employed for dataset collection, resulting in frames with diverse moiré patterns. 
In addition, we utilize the raw image demoiréing dataset TMM22~\cite{yue2022recaptured} to verify the generalization ability of our method. TMM22 comprises 540 paired raw and sRGB images with ground truth for training and 408 pairs for testing, covering various recaptured scenes such as natural images, webpages, and documents. To facilitate the training and comparison process, patches sized 256$\times$256 and 512$\times$512 are cropped for the training and testing sets, respectively.

\noindent \textbf{Training Details.}
DemMamba is trained using the AdamW optimizer, with $\beta_1$ set to 0.9 and $\beta_2$ set to 0.999. We employ a multistep learning rate schedule, with the learning rate initialized at $4 \times 10^{-4}$. DemMamba is trained for 100 epochs using a batch size of 8 on 4 NVIDIA Tesla A800 GPUs.
For the structure in DemMamba, the numbers of FASTMG, SMB, and TMB are all set to 4. The weighting coefficient in SMB($\alpha_1$) and TMB($\alpha_2$) are both set to 0.5.

\subsection{Quantitative Results}
\textbf{Frame Level Comparison.}
We compare our approach with two raw image demoiréing methods, RDNet~\cite{yue2022recaptured} and RRID~\cite{xu2024image}, along with several video demoiréing methods: VDmoiré~\cite{dai2022video}, DTNet~\cite{xu2024direction} and RawVD~\cite{cheng2024recaptured}. Since VDmoiré and DTNet are originally designed for video demoiréing in the sRGB domain, we have retrained them with raw inputs, adjusting the input channel number and inserting upsampling layers to adapt them to raw data. The revised versions are marked with an $\ast$.
To evaluate the performance of our proposed DemMamba, we adopt the following three standard metrics to assess pixel-wise accuracy and the perceptual quality: PSNR, SSIM~\cite{wang2004image}, and LPIPS~\cite{zhang2018unreasonable}. 
Additionally, we assess model complexity by analyzing the number of parameters and inference time. To ensure a fair comparison, all models are fine-tuned using the default settings provided in their respective papers. We select the better results from the pretrained and retrained models for comparison, providing an advantage over competing methods.

\begin{figure*}[!htbp]
\centering
\includegraphics[width=1\linewidth]{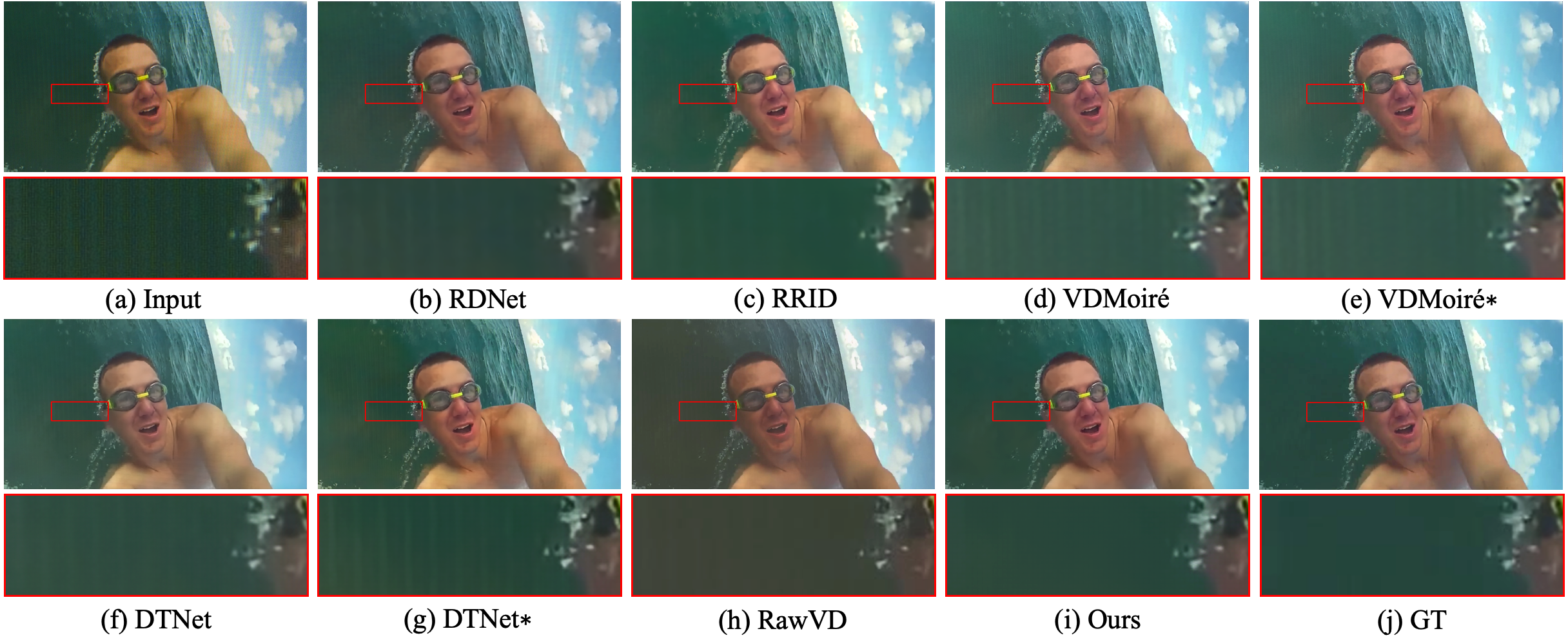}
\caption{Qualitative comparison on raw video demoiréing RawVDmoiré dataset.}
\label{fig:nips23}
\end{figure*}

Table~\ref{table:nips23} presents the quantitative comparison and computational complexity analysis on the RawVDmoiré dataset.
Our proposed method, DemMamba, surpasses the second-best approach, RawVD, achieving a PSNR of 30.069 dB with an inference time of only 0.686 seconds, demonstrating both efficiency and effectiveness in demoiréing.
Furthermore, DemMamba maintains this performance with a modest parameter count of 2.941M.
Additionally, we provide a larger version of our model, denoted as Ours-L, achieves an additional 1.6 dB improvement in PSNR over RawVD. Improvements are also observed in SSIM and LPIPS metrics, further substantiating the superiority of our approach.

\begin{figure*}[!htbp]
\centering
\includegraphics[width=0.95\linewidth]{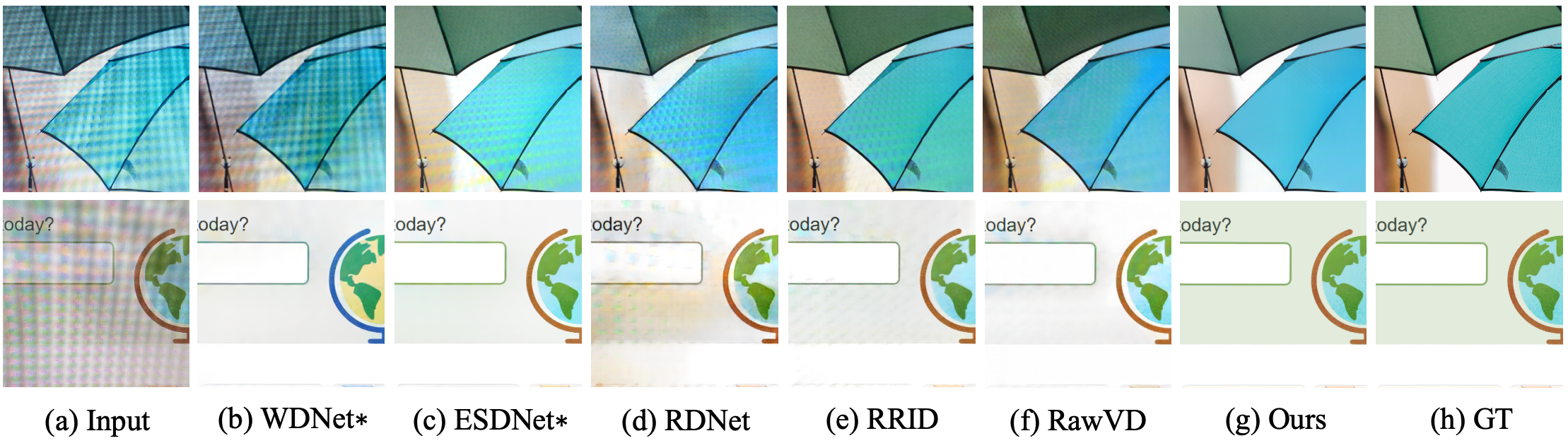}
\caption{Qualitative comparison on raw image demoiréing TMM22 dataset.}
\label{fig:tmm22}
\end{figure*}

\begin{table}[htbp]
    \begin{center}
    \footnotesize
    
        \begin{tabular}{cccccc}
        \toprule
        \textbf{Method} & \textbf{PSNR$\uparrow$} & \textbf{SSIM$\uparrow$} & \textbf{LPIPS$\downarrow$} & \textbf{Runtime (s)}\\
        \midrule 
            DMCNN & 23.54 & 0.885 & 0.154 & \textbf{0.052}\\
            DMCNN$\ast$ & 24.51 & 0.894 & 0.138 & 0.058 \\
            WDNet & 22.33 & 0.802 & 0.166 & 0.284 \\
            WDNet$\ast$  & 23.08 & 0.815 & 0.149 & 0.329 \\
            ESDNet & 26.77 & 0.927 & 0.089 & \underline{0.115} \\
            ESDNet$\ast$ & \underline{27.48} & 0.934 & 0.078 & 0.129\\
            RDNet & 26.16 & 0.921 & 0.091 & 1.094  \\
            RRID & 27.24 & 0.929 & 0.098 & 0.326\\
            RawVD & 27.26 & \underline{0.935} & \underline{0.075} & 0.182 \\
            Ours & \textbf{28.14} & \textbf{0.936} & \textbf{0.067} & 0.159 \\
        \bottomrule
    \end{tabular}
    \caption{Quantitative comparison on TMM22 dataset.}
    \label{table:tmm22}
\end{center}
\end{table}

To verify the generalization of our method, we conduct additional experiments on the raw image demoiréing dataset TMM22~\cite{yue2022recaptured}. To adapt to the TMM22 dataset, we input the same three frames into our model. In Table~\ref{table:tmm22}, we compare our method with several image demoiréing methods, including DMCNN~\cite{sun2018moire}, WDNet~\cite{liu2020wavelet}, and ESDNet~\cite{yu2022towards} (along with their corresponding raw input versions, denoted with $\ast$), as well as raw image demoiréing methods RDNet and RRID, and the raw video demoiréing method RawVD. Although DemMamba is initially designed for raw video demoiréing, it achieves the best performance on TMM22 dataset with a PSNR of 28.14 dB and an SSIM of 0.936.
Regarding perceptual quality, DemMamba surpasses all the other methods with an LPIPS score of 0.067.
The qualitative results on the TMM22 dataset demonstrate the generalization capability of our method.

\noindent
\textbf{Video-Level Comparison.}
We utilize FVD ~\cite{unterthiner2018towards} and FSIM ~\cite{zhang2011fsim} to assess the quality of video outputs. FVD, which extends the Fréchet Inception Distance concept, evaluates the temporal coherence of videos. 
FSIM is an image quality assessment index that evaluates changes in image quality by comparing the similarity of low-level content features, such as gradients and phase consistency.
As indicated in Table~\ref{table:nips23}, DemMamba outperforms existing approaches in both metrics, demonstrating that our outputs more closely resemble the target distribution of entire video sequences and better preserve per-pixel and structural visual information. Additionally, we have included a video file in the supplementary material to provide a visual comparison of video quality.

\subsection{Qualitative Results}
Visual comparisons between the proposed method and existing methods are presented in Fig.~\ref{fig:nips23}. The results demonstrate DemMamba's effectiveness in removing moiré patterns and correcting color deviations. Notably, vertical moiré stripes are visible in the background regions. Images restored by RDNet, RRID, VDmoiré, and DTNet still contain residual moiré artifacts. Although RawVD can eliminate most moiré patterns, the restored images exhibit a slight grayish bias. In contrast, DemMamba effectively removes moiré patterns while accurately restoring colors.

Additionally, we present the qualitative comparison on raw image demoiréing dataset TMM22 in Fig.~\ref{fig:tmm22}.
In the first scene, which features a blue umbrella with moiré patterns, previous methods tend to leave residual moiré patterns in their restored images.
In the second scene, we present a case of webpages affected by moiré patterns. Although some methods partially remove moiré patterns, they struggle to restore the original colors accurately. DemMamba's ability to model long-range dependencies enables a more accurate restoration of colors, resulting in enhanced visual quality.

\subsection{Ablation Study}
\begin{figure}[!ht]
\centering
\includegraphics[width=0.7\linewidth]{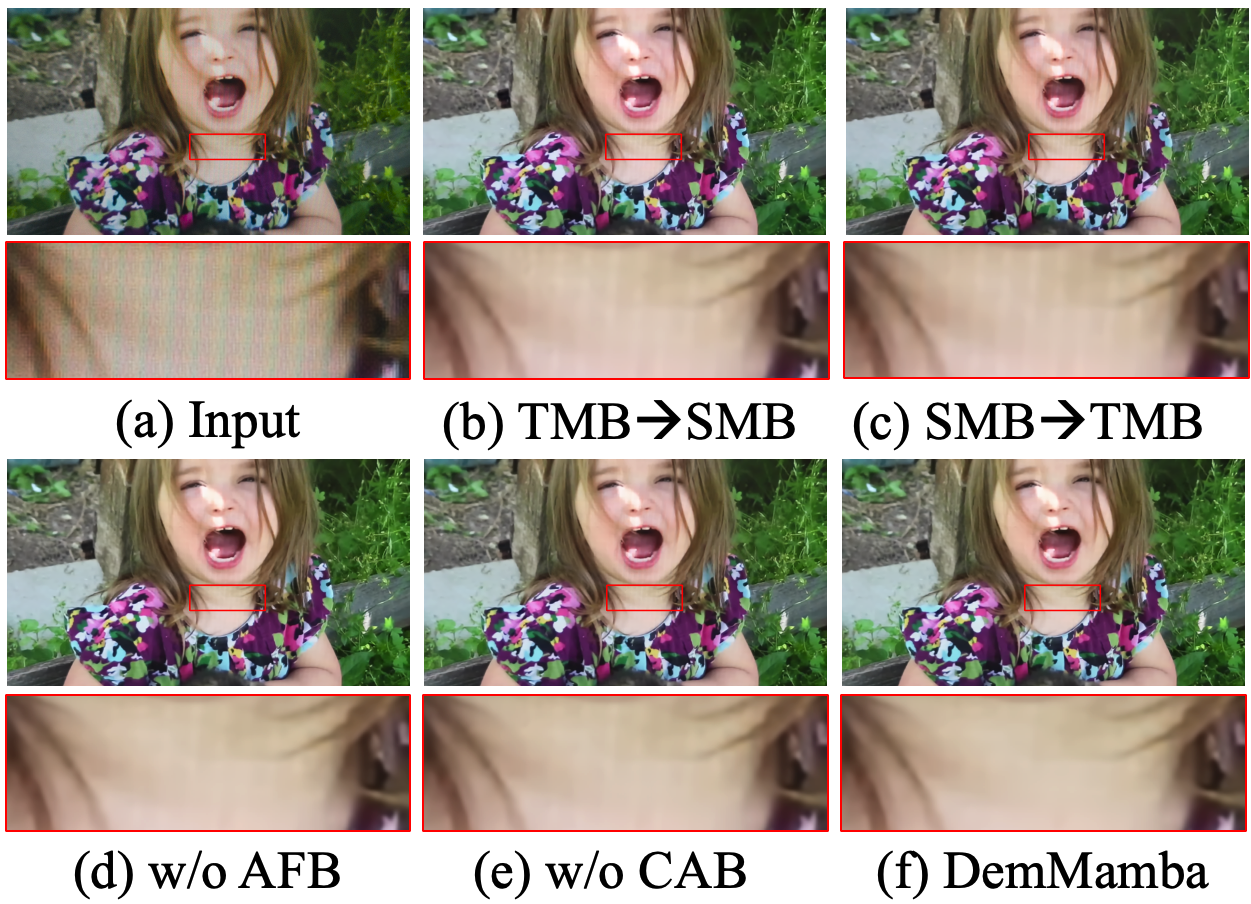}
\caption{Different design choices of FASTMG.}
\label{fig:ablation_p1}
\end{figure}

\textbf{Different design choices of FASTMG.} Table~\ref{table:ablation_FASTMG} shows an assessment of DemMamba through ablation experiments with various combinations of the foundational elements.
Replacing TMB with SMB results in a PSNR of 30.15 dB and an SSIM of 0.9187, whereas the opposite substitution (replace SMB with TMB) further degrades performance to 29.92 dB and 0.9179, confirming that neither block alone can fully capture both spatial and temporal correlations. 
Furthermore, removing AFB and CAB respectively leads to slight performance decreases, underscoring the importance of the two modules in moire pattern removal and temporal information utilization. 
In Fig.~\ref{fig:ablation_p1}, we also provide visual representations to further elucidate the effectiveness of the proposed modules in removing moiré patterns.
\begin{table}[htbp]
    \begin{center}
    \footnotesize
        \begin{tabular}{ccc}
        \toprule
        \textbf{Models} & \textbf{PSNR$\uparrow$} & \textbf{SSIM$\uparrow$}\\
        \midrule 
            replace TMB with SMB & 30.15 & 0.9187  \\
            replace SMB with TMB & 29.92 & 0.9179\\
            remove AFB & 30.22 & 0.9192 \\
            remove CAB & 30.18 & 0.9190 \\
            DemMamba (Ours) & \textbf{30.32} & \textbf{0.9195}\\
        \bottomrule
    \end{tabular}
    \caption{Different design choices of FASTMG.}
    \label{table:ablation_FASTMG}
\end{center}
\end{table}

\begin{figure}[!ht]
\centering
\includegraphics[width=1\linewidth]{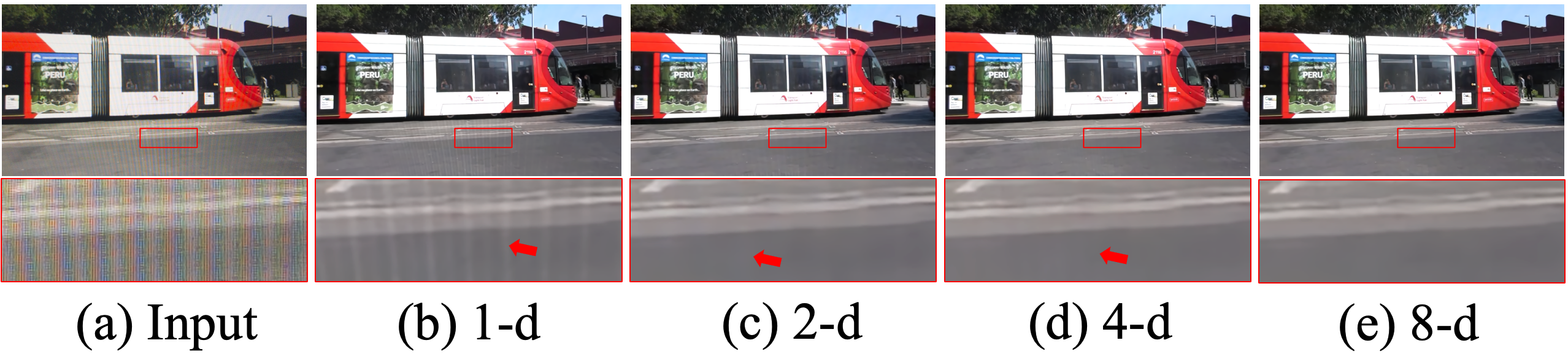}
\caption{Different scan modes in SMB.}
\label{fig:ablation_p2}
\end{figure}

\noindent\textbf{Different scan modes in SMB.}
Table~\ref{table:ablation_scan} reports the performance of the SMB when employing varying numbers of scan modes. Increasing the number of scan directions from one (1-d) to eight (8-d) yields improvements of 0.57 dB in PSNR and 0.0026 in SSIM, with acceptable increases in parameter count and runtime. Moreover, Fig.~\ref{fig:ablation_p2} visually demonstrates that increasing the number of scan directions progressively reduces moiré artifacts across multiple orientations, further substantiating the effectiveness of the proposed multi-directional scanning strategy.

\begin{table}[htbp]
    \begin{center}
    \footnotesize
        \begin{tabular}{ccccc}
        \toprule
        \textbf{Scan mode} & \textbf{PSNR$\uparrow$} & \textbf{SSIM$\uparrow$} & \textbf{Params(M)} & \textbf{Runtime(s)}\\
        \midrule 
            1-d & 29.75 & 0.9169 & 3.797 & 0.204\\
            2-d & 29.98 & 0.9182 & 3.881 & 0.221 \\
            4-d & 30.24 & 0.9191 & 4.048 & 0.454 \\
            8-d & 30.32 & 0.9195 & 4.382 & 0.686 \\
        \bottomrule
    \end{tabular}
    \caption{Different scan modes in SMB.}
    \label{table:ablation_scan}
\end{center}
\end{table}

\section{Conclusion}
We propose DemMamba, an alignment-free raw video demoiréing method enhanced by frequency-assisted spatio-temporal Mamba. DemMamba introduces SMB and TMB to model the inter- and intra-relationships, mitigating moiré patterns while ensuring temporal consistency.
The SMB employs a multi-directional scanning mechanism coupled with a learnable frequency compressor to differentiate interference patterns across various orientations and frequencies.
TMB enhances the utilization of adjacent data to maintain temporal stability, supplemented by a CAB that strengthens temporal information integration.
Extensive experiments demonstrate that DemMamba outperforms state-of-the-art methods both qualitatively and quantitatively.

\bibliography{aaai2026}

\begin{thebibliography}{62}
\providecommand{\natexlab}[1]{#1}

\bibitem[{Bai et~al.(2024)Bai, Yin, He, Li, and Zhang}]{bai2024retinexmamba}
Bai, J.; Yin, Y.; He, Q.; Li, Y.; and Zhang, X. 2024.
\newblock Retinexmamba: Retinex-based mamba for low-light image enhancement.
\newblock In \emph{International Conference on Neural Information Processing},
  427--442. Springer.

\bibitem[{Behrouz, Santacatterina, and Zabih(2024)}]{behrouz2024mambamixer}
Behrouz, A.; Santacatterina, M.; and Zabih, R. 2024.
\newblock Mambamixer: Efficient selective state space models with dual token
  and channel selection.
\newblock \emph{arXiv preprint arXiv:2403.19888}.

\bibitem[{Chen et~al.(2024)Chen, Huang, Xu, Pei, Chen, Li, Wang, Li, Lu, and
  Wang}]{chen2024video}
Chen, G.; Huang, Y.; Xu, J.; Pei, B.; Chen, Z.; Li, Z.; Wang, J.; Li, K.; Lu,
  T.; and Wang, L. 2024.
\newblock Video mamba suite: State space model as a versatile alternative for
  video understanding.
\newblock \emph{arXiv preprint arXiv:2403.09626}.

\bibitem[{Cheng, Fu, and Yang(2019)}]{cheng2019multi}
Cheng, X.; Fu, Z.; and Yang, J. 2019.
\newblock Multi-scale dynamic feature encoding network for image
  demoir{\'e}ing.
\newblock In \emph{2019 IEEE/CVF International Conference on Computer Vision
  Workshop (ICCVW)}, 3486--3493. IEEE.

\bibitem[{Cheng, Liu, and Yang(2024)}]{cheng2024recaptured}
Cheng, Y.; Liu, X.; and Yang, J. 2024.
\newblock Recaptured Raw Screen Image and Video Demoir{\'e}ing via Channel and
  Spatial Modulations.
\newblock \emph{Advances in Neural Information Processing Systems}, 36.

\bibitem[{Dai et~al.(2022)Dai, Yu, Ma, Zhang, Li, Li, Shen, and
  Qi}]{dai2022video}
Dai, P.; Yu, X.; Ma, L.; Zhang, B.; Li, J.; Li, W.; Shen, J.; and Qi, X. 2022.
\newblock Video demoireing with relation-based temporal consistency.
\newblock In \emph{Proceedings of the IEEE/CVF Conference on Computer Vision
  and Pattern Recognition}, 17622--17631.

\bibitem[{Deng and Gu(2024)}]{deng2024cu}
Deng, R.; and Gu, T. 2024.
\newblock Cu-mamba: Selective state space models with channel learning for
  image restoration.
\newblock \emph{arXiv preprint arXiv:2404.11778}.

\bibitem[{Dong et~al.(2022)Dong, Xu, Miao, Ma, Zhang, Yang, Jin, Teoh, and
  Shen}]{dong2022abandoning}
Dong, X.; Xu, W.; Miao, Z.; Ma, L.; Zhang, C.; Yang, J.; Jin, Z.; Teoh, A.
  B.~J.; and Shen, J. 2022.
\newblock Abandoning the bayer-filter to see in the dark.
\newblock In \emph{Proceedings of the IEEE/CVF Conference on Computer Vision
  and Pattern Recognition}, 17431--17440.

\bibitem[{Gu and Dao(2023)}]{gu2023mamba}
Gu, A.; and Dao, T. 2023.
\newblock Mamba: Linear-time sequence modeling with selective state spaces.
\newblock \emph{arXiv preprint arXiv:2312.00752}.

\bibitem[{Gu, Goel, and R{\'e}(2021)}]{gu2021efficiently}
Gu, A.; Goel, K.; and R{\'e}, C. 2021.
\newblock Efficiently modeling long sequences with structured state spaces.
\newblock \emph{arXiv preprint arXiv:2111.00396}.

\bibitem[{Gu et~al.(2021)Gu, Johnson, Goel, Saab, Dao, Rudra, and
  R{\'e}}]{gu2021combining}
Gu, A.; Johnson, I.; Goel, K.; Saab, K.; Dao, T.; Rudra, A.; and R{\'e}, C.
  2021.
\newblock Combining recurrent, convolutional, and continuous-time models with
  linear state space layers.
\newblock \emph{Advances in neural information processing systems}, 34:
  572--585.

\bibitem[{Guo et~al.(2025)Guo, Li, Dai, Ouyang, Ren, and Xia}]{guo2025mambair}
Guo, H.; Li, J.; Dai, T.; Ouyang, Z.; Ren, X.; and Xia, S.-T. 2025.
\newblock Mambair: A simple baseline for image restoration with state-space
  model.
\newblock In \emph{European Conference on Computer Vision}, 222--241. Springer.

\bibitem[{He et~al.(2019)He, Wang, Shi, and Duan}]{he2019mop}
He, B.; Wang, C.; Shi, B.; and Duan, L.-Y. 2019.
\newblock Mop Moire Patterns Using MopNet.
\newblock 2424--2432.

\bibitem[{He et~al.(2020)He, Wang, Shi, and Duan}]{he2020fhde}
He, B.; Wang, C.; Shi, B.; and Duan, L.-Y. 2020.
\newblock Fhde 2 net: Full high definition demoireing network.
\newblock In \emph{Computer Vision--ECCV 2020: 16th European Conference,
  Glasgow, UK, August 23--28, 2020, Proceedings, Part XXII 16}, 713--729.
  Springer.

\bibitem[{Huang et~al.(2022)Huang, Yang, Hu, Liu, and Duan}]{huang2022towards}
Huang, H.; Yang, W.; Hu, Y.; Liu, J.; and Duan, L.-Y. 2022.
\newblock Towards low light enhancement with raw images.
\newblock \emph{IEEE Transactions on Image Processing}, 31: 1391--1405.

\bibitem[{Jin et~al.(2023)Jin, Han, Li, Guo, Chai, and Li}]{jin2023dnf}
Jin, X.; Han, L.-H.; Li, Z.; Guo, C.-L.; Chai, Z.; and Li, C. 2023.
\newblock DNF: Decouple and Feedback Network for Seeing in the Dark.
\newblock In \emph{Proceedings of the IEEE/CVF Conference on Computer Vision
  and Pattern Recognition}, 18135--18144.

\bibitem[{Johnson, Alahi, and Fei-Fei(2016)}]{johnson2016perceptual}
Johnson, J.; Alahi, A.; and Fei-Fei, L. 2016.
\newblock Perceptual losses for real-time style transfer and super-resolution.
\newblock In \emph{Computer Vision--ECCV 2016: 14th European Conference,
  Amsterdam, The Netherlands, October 11-14, 2016, Proceedings, Part II 14},
  694--711. Springer.

\bibitem[{Kalman(1960)}]{kalman1960new}
Kalman, R.~E. 1960.
\newblock A new approach to linear filtering and prediction problems.

\bibitem[{Kong et~al.(2024)Kong, Dong, Yang, and Pan}]{kong2024efficient}
Kong, L.; Dong, J.; Yang, M.-H.; and Pan, J. 2024.
\newblock Efficient Visual State Space Model for Image Deblurring.
\newblock \emph{arXiv preprint arXiv:2405.14343}.

\bibitem[{Lei and Chen(2021)}]{lei2021robust}
Lei, C.; and Chen, Q. 2021.
\newblock Robust reflection removal with reflection-free flash-only cues.
\newblock In \emph{Proceedings of the IEEE/CVF Conference on Computer Vision
  and Pattern Recognition}, 14811--14820.

\bibitem[{Li et~al.(2024)Li, Liu, Fu, Xu, and Zha}]{li2024fouriermamba}
Li, D.; Liu, Y.; Fu, X.; Xu, S.; and Zha, Z.-J. 2024.
\newblock FourierMamba: Fourier Learning Integration with State Space Models
  for Image Deraining.
\newblock \emph{arXiv preprint arXiv:2405.19450}.

\bibitem[{Lin et~al.(2024)Lin, Lin, Chen, and Hua}]{lin2024pixmamba}
Lin, W.-T.; Lin, Y.-X.; Chen, J.-W.; and Hua, K.-L. 2024.
\newblock PixMamba: Leveraging State Space Models in a Dual-Level Architecture
  for Underwater Image Enhancement.
\newblock \emph{arXiv preprint arXiv:2406.08444}.

\bibitem[{Liu, Shu, and Wu(2018)}]{liu2018demoir}
Liu, B.; Shu, X.; and Wu, X. 2018.
\newblock Demoir$\backslash$'eing of Camera-Captured Screen Images Using Deep
  Convolutional Neural Network.
\newblock \emph{arXiv preprint arXiv:1804.03809}.

\bibitem[{Liu et~al.(2024{\natexlab{a}})Liu, An, Yuan, Zhou, Li, Wang, and
  Tian}]{liu2024video}
Liu, L.; An, J.; Yuan, S.; Zhou, W.; Li, H.; Wang, Y.; and Tian, Q.
  2024{\natexlab{a}}.
\newblock Video Demoir{\'e}ing with Deep Temporal Color Embedding and
  Video-Image Invertible Consistency.
\newblock \emph{IEEE Transactions on Multimedia}.

\bibitem[{Liu et~al.(2020{\natexlab{a}})Liu, Liu, Yuan, Slabaugh, Leonardis,
  Zhou, and Tian}]{liu2020wavelet}
Liu, L.; Liu, J.; Yuan, S.; Slabaugh, G.; Leonardis, A.; Zhou, W.; and Tian, Q.
  2020{\natexlab{a}}.
\newblock Wavelet-based dual-branch network for image demoir{\'e}ing.
\newblock In \emph{Computer Vision--ECCV 2020: 16th European Conference,
  Glasgow, UK, August 23--28, 2020, Proceedings, Part XIII 16}, 86--102.
  Springer.

\bibitem[{Liu et~al.(2020{\natexlab{b}})Liu, Yuan, Liu, Bao, Slabaugh, and
  Tian}]{liu2020self}
Liu, L.; Yuan, S.; Liu, J.; Bao, L.; Slabaugh, G.; and Tian, Q.
  2020{\natexlab{b}}.
\newblock Self-adaptively learning to demoir{\'e} from focused and defocused
  image pairs.
\newblock \emph{Advances in Neural Information Processing Systems}, 33:
  22282--22292.

\bibitem[{Liu et~al.(2024{\natexlab{b}})Liu, Dan, Lu, Yu, Li, and
  Li}]{liu2024cm}
Liu, M.; Dan, J.; Lu, Z.; Yu, Y.; Li, Y.; and Li, X. 2024{\natexlab{b}}.
\newblock CM-UNet: Hybrid CNN-Mamba UNet for Remote Sensing Image Semantic
  Segmentation.
\newblock \emph{arXiv preprint arXiv:2405.10530}.

\bibitem[{Liu et~al.(2020{\natexlab{c}})Liu, Li, Nan, Zong, and
  Song}]{liu2020mmdm}
Liu, S.; Li, C.; Nan, N.; Zong, Z.; and Song, R. 2020{\natexlab{c}}.
\newblock MMDM: Multi-frame and multi-scale for image demoir{\'e}ing.
\newblock In \emph{Proceedings of the IEEE/CVF Conference on Computer Vision
  and Pattern Recognition Workshops}, 434--435.

\bibitem[{Liu et~al.(2024{\natexlab{c}})Liu, Tian, Zhao, Yu, Xie, Wang, Ye,
  Jiao, and Liu}]{liu2024vmamba}
Liu, Y.; Tian, Y.; Zhao, Y.; Yu, H.; Xie, L.; Wang, Y.; Ye, Q.; Jiao, J.; and
  Liu, Y. 2024{\natexlab{c}}.
\newblock Vmamba: Visual state space model.
\newblock In \emph{The Thirty-eighth Annual Conference on Neural Information
  Processing Systems}.

\bibitem[{Luo, Wu, and Guo(2024)}]{luo2024and}
Luo, F.; Wu, X.; and Guo, Y. 2024.
\newblock AND: Adversarial neural degradation for learning blind image
  super-resolution.
\newblock \emph{Advances in Neural Information Processing Systems}, 36.

\bibitem[{Niu et~al.(2023)Niu, Lin, Liu, and Guo}]{niu2023progressive}
Niu, Y.; Lin, Z.; Liu, W.; and Guo, W. 2023.
\newblock Progressive Moire Removal and Texture Complementation for Image
  Demoireing.
\newblock \emph{IEEE Transactions on Circuits and Systems for Video
  Technology}.

\bibitem[{Niu et~al.(2024)Niu, Xu, Lin, and Liu}]{niu2024std}
Niu, Y.; Xu, R.; Lin, Z.; and Liu, W. 2024.
\newblock STD-Net: Spatio-Temporal Decomposition Network for Video
  Demoir{\'e}ing with Sparse Transformers.
\newblock \emph{IEEE Transactions on Circuits and Systems for Video
  Technology}.

\bibitem[{Patro and Agneeswaran(2024)}]{patro2024simba}
Patro, B.~N.; and Agneeswaran, V.~S. 2024.
\newblock Simba: Simplified mamba-based architecture for vision and
  multivariate time series.
\newblock \emph{arXiv preprint arXiv:2403.15360}.

\bibitem[{Quan et~al.(2023)Quan, Huang, He, and Xu}]{quan2023deep}
Quan, Y.; Huang, H.; He, S.; and Xu, R. 2023.
\newblock Deep Video Demoir{\'e}ing via Compact Invertible Dyadic
  Decomposition.
\newblock In \emph{Proceedings of the IEEE/CVF International Conference on
  Computer Vision}, 12677--12686.

\bibitem[{Shi et~al.(2024)Shi, Xia, Jin, Wang, Zhao, Xia, Xiao, and
  Yang}]{shi2024vmambair}
Shi, Y.; Xia, B.; Jin, X.; Wang, X.; Zhao, T.; Xia, X.; Xiao, X.; and Yang, W.
  2024.
\newblock Vmambair: Visual state space model for image restoration.
\newblock \emph{arXiv preprint arXiv:2403.11423}.

\bibitem[{Song et~al.(2023)Song, Zhou, Chen, and Zhang}]{song2023real}
Song, B.; Zhou, J.; Chen, X.; and Zhang, S. 2023.
\newblock Real-Scene Reflection Removal with RAW-RGB Image Pairs.
\newblock \emph{IEEE Transactions on Circuits and Systems for Video
  Technology}.

\bibitem[{Sun, Yu, and Wang(2018)}]{sun2018moire}
Sun, Y.; Yu, Y.; and Wang, W. 2018.
\newblock Moir{\'e} photo restoration using multiresolution convolutional
  neural networks.
\newblock \emph{IEEE Transactions on Image Processing}, 27(8): 4160--4172.

\bibitem[{Unterthiner et~al.(2018)Unterthiner, Van~Steenkiste, Kurach,
  Marinier, Michalski, and Gelly}]{unterthiner2018towards}
Unterthiner, T.; Van~Steenkiste, S.; Kurach, K.; Marinier, R.; Michalski, M.;
  and Gelly, S. 2018.
\newblock Towards accurate generative models of video: A new metric \&
  challenges.
\newblock \emph{arXiv preprint arXiv:1812.01717}.

\bibitem[{Wan et~al.(2024)Wan, Wang, Yong, Zhang, Stepputtis, Sycara, and
  Xie}]{wan2024sigma}
Wan, Z.; Wang, Y.; Yong, S.; Zhang, P.; Stepputtis, S.; Sycara, K.; and Xie, Y.
  2024.
\newblock Sigma: Siamese mamba network for multi-modal semantic segmentation.
\newblock \emph{arXiv preprint arXiv:2404.04256}.

\bibitem[{Wang et~al.(2023{\natexlab{a}})Wang, He, Wu, Wan, Shi, and
  Duan}]{wang2023coarse}
Wang, C.; He, B.; Wu, S.; Wan, R.; Shi, B.; and Duan, L.-Y. 2023{\natexlab{a}}.
\newblock Coarse-to-fine Disentangling Demoir{\'e}ing Framework for Recaptured
  Screen Images.
\newblock \emph{IEEE Transactions on Pattern Analysis and Machine
  Intelligence}.

\bibitem[{Wang et~al.(2021)Wang, Tian, Li, and Guo}]{wang2021image}
Wang, H.; Tian, Q.; Li, L.; and Guo, X. 2021.
\newblock Image demoir{\'e}ing with a dual-domain distilling network.
\newblock In \emph{2021 IEEE International Conference on Multimedia and Expo
  (ICME)}, 1--6. IEEE.

\bibitem[{Wang et~al.(2023{\natexlab{b}})Wang, Zhu, Wang, Yu, Liu, Omar, and
  Hamid}]{wang2023selective}
Wang, J.; Zhu, W.; Wang, P.; Yu, X.; Liu, L.; Omar, M.; and Hamid, R.
  2023{\natexlab{b}}.
\newblock Selective structured state-spaces for long-form video understanding.
\newblock In \emph{Proceedings of the IEEE/CVF Conference on Computer Vision
  and Pattern Recognition}, 6387--6397.

\bibitem[{Wang et~al.(2019)Wang, Chan, Yu, Dong, and Change~Loy}]{wang2019edvr}
Wang, X.; Chan, K.~C.; Yu, K.; Dong, C.; and Change~Loy, C. 2019.
\newblock Edvr: Video restoration with enhanced deformable convolutional
  networks.
\newblock In \emph{Proceedings of the IEEE/CVF conference on computer vision
  and pattern recognition workshops}, 0--0.

\bibitem[{Wang et~al.(2004)Wang, Bovik, Sheikh, and Simoncelli}]{wang2004image}
Wang, Z.; Bovik, A.~C.; Sheikh, H.~R.; and Simoncelli, E.~P. 2004.
\newblock Image quality assessment: from error visibility to structural
  similarity.
\newblock \emph{IEEE transactions on image processing}, 13(4): 600--612.

\bibitem[{Wu et~al.(2024)Wu, Yang, Xu, Wang, Zhou, and Zhu}]{wu2024rainmamba}
Wu, H.; Yang, Y.; Xu, H.; Wang, W.; Zhou, J.; and Zhu, L. 2024.
\newblock RainMamba: Enhanced Locality Learning with State Space Models for
  Video Deraining.
\newblock \emph{arXiv preprint arXiv:2407.21773}.

\bibitem[{Xiao, Lu, and Wang(2020)}]{xiao2024p}
Xiao, Z.; Lu, Z.; and Wang, X. 2020.
\newblock P-BiC: Ultra-High-Definition Image Demoireing via Patch Bilateral
  Compensation.
\newblock In \emph{ACM Multimedia 2024}.

\bibitem[{Xing and Egiazarian(2021)}]{xing2021end}
Xing, W.; and Egiazarian, K. 2021.
\newblock End-to-end learning for joint image demosaicing, denoising and
  super-resolution.
\newblock In \emph{Proceedings of the IEEE/CVF conference on computer vision
  and pattern recognition}, 3507--3516.

\bibitem[{Xu et~al.(2024{\natexlab{a}})Xu, Song, Chen, Liu, and
  Zhou}]{xu2024image}
Xu, S.; Song, B.; Chen, X.; Liu, X.; and Zhou, J. 2024{\natexlab{a}}.
\newblock Image demoireing in raw and srgb domains.
\newblock In \emph{European Conference on Computer Vision}, 108--124. Springer.

\bibitem[{Xu et~al.(2024{\natexlab{b}})Xu, Song, Chen, and
  Zhou}]{xu2024direction}
Xu, S.; Song, B.; Chen, X.; and Zhou, J. 2024{\natexlab{b}}.
\newblock Direction-aware video demoireing with temporal-guided bilateral
  learning.
\newblock In \emph{Proceedings of the AAAI Conference on Artificial
  Intelligence}, volume~38, 6360--6368.

\bibitem[{Yang et~al.(2024)Yang, Chen, Espinosa, Ericsson, Wang, Liu, and
  Crowley}]{yang2024plainmamba}
Yang, C.; Chen, Z.; Espinosa, M.; Ericsson, L.; Wang, Z.; Liu, J.; and Crowley,
  E.~J. 2024.
\newblock Plainmamba: Improving non-hierarchical mamba in visual recognition.
\newblock \emph{arXiv preprint arXiv:2403.17695}.

\bibitem[{Yang et~al.(2025)Yang, Zhang, Jin, Cheng, Jiang, Yue, and
  Yang}]{yang2025dsdnet}
Yang, Q.; Zhang, F.; Jin, Y.; Cheng, Q.; Jiang, P.; Yue, H.; and Yang, J. 2025.
\newblock DSDNet: Raw Domain Demoir$\backslash$'eing via Dual Color-Space
  Synergy.
\newblock \emph{arXiv preprint arXiv:2504.15756}.

\bibitem[{Yu et~al.(2022)Yu, Dai, Li, Ma, Shen, Li, and Qi}]{yu2022towards}
Yu, X.; Dai, P.; Li, W.; Ma, L.; Shen, J.; Li, J.; and Qi, X. 2022.
\newblock Towards efficient and scale-robust ultra-high-definition image
  demoir{\'e}ing.
\newblock In \emph{Computer Vision--ECCV 2022: 17th European Conference, Tel
  Aviv, Israel, October 23--27, 2022, Proceedings, Part XVIII}, 646--662.
  Springer.

\bibitem[{Yue et~al.(2022)Yue, Cheng, Mao, Cao, and Yang}]{yue2022recaptured}
Yue, H.; Cheng, Y.; Mao, Y.; Cao, C.; and Yang, J. 2022.
\newblock Recaptured screen image demoir{\'e}ing in raw domain.
\newblock \emph{IEEE Transactions on Multimedia}.

\bibitem[{Yue, Zhang, and Yang(2022)}]{yue2022real}
Yue, H.; Zhang, Z.; and Yang, J. 2022.
\newblock Real-RawVSR: Real-World Raw Video Super-Resolution with a Benchmark
  Dataset.
\newblock In \emph{European Conference on Computer Vision}, 608--624. Springer.

\bibitem[{Zhang et~al.(2011)Zhang, Zhang, Mou, and Zhang}]{zhang2011fsim}
Zhang, L.; Zhang, L.; Mou, X.; and Zhang, D. 2011.
\newblock FSIM: A feature similarity index for image quality assessment.
\newblock \emph{IEEE transactions on Image Processing}, 20(8): 2378--2386.

\bibitem[{Zhang et~al.(2018{\natexlab{a}})Zhang, Isola, Efros, Shechtman, and
  Wang}]{zhang2018unreasonable}
Zhang, R.; Isola, P.; Efros, A.~A.; Shechtman, E.; and Wang, O.
  2018{\natexlab{a}}.
\newblock The unreasonable effectiveness of deep features as a perceptual
  metric.
\newblock In \emph{Proceedings of the IEEE conference on computer vision and
  pattern recognition}, 586--595.

\bibitem[{Zhang et~al.(2018{\natexlab{b}})Zhang, Li, Li, Wang, Zhong, and
  Fu}]{zhang2018image}
Zhang, Y.; Li, K.; Li, K.; Wang, L.; Zhong, B.; and Fu, Y. 2018{\natexlab{b}}.
\newblock Image super-resolution using very deep residual channel attention
  networks.
\newblock In \emph{Proceedings of the European conference on computer vision
  (ECCV)}, 286--301.

\bibitem[{Zhang et~al.(2023)Zhang, Lin, Li, Liu, Wang, Chao, Ren, Wen, Chen,
  and Ji}]{zhang2023real}
Zhang, Y.; Lin, M.; Li, X.; Liu, H.; Wang, G.; Chao, F.; Ren, S.; Wen, Y.;
  Chen, X.; and Ji, R. 2023.
\newblock Real-Time Image Demoireing on Mobile Devices.
\newblock \emph{arXiv preprint arXiv:2302.02184}.

\bibitem[{Zhen, Hu, and Feng(2024)}]{zhen2024freqmamba}
Zhen, Z.; Hu, Y.; and Feng, Z. 2024.
\newblock Freqmamba: Viewing mamba from a frequency perspective for image
  deraining.
\newblock \emph{arXiv preprint arXiv:2404.09476}.

\bibitem[{Zheng et~al.(2020)Zheng, Yuan, Slabaugh, and
  Leonardis}]{zheng2020image}
Zheng, B.; Yuan, S.; Slabaugh, G.; and Leonardis, A. 2020.
\newblock Image demoireing with learnable bandpass filters.
\newblock In \emph{Proceedings of the IEEE/CVF Conference on Computer Vision
  and Pattern Recognition}, 3636--3645.

\bibitem[{Zhu et~al.(2024{\natexlab{a}})Zhu, Liao, Zhang, Wang, Liu, and
  Wang}]{zhu2024vision}
Zhu, L.; Liao, B.; Zhang, Q.; Wang, X.; Liu, W.; and Wang, X.
  2024{\natexlab{a}}.
\newblock Vision mamba: Efficient visual representation learning with
  bidirectional state space model.
\newblock \emph{arXiv preprint arXiv:2401.09417}.

\bibitem[{Zhu et~al.(2024{\natexlab{b}})Zhu, Fang, Cai, Chen, and
  Fan}]{zhu2024rethinking}
Zhu, Q.; Fang, Y.; Cai, Y.; Chen, C.; and Fan, L. 2024{\natexlab{b}}.
\newblock Rethinking Scanning Strategies with Vision Mamba in Semantic
  Segmentation of Remote Sensing Imagery: An Experimental Study.
\newblock \emph{arXiv preprint arXiv:2405.08493}.

\end{thebibliography}

\end{document}